# Design Of Fuzzy Logic Traffic Controller For Isolated Intersections With Emergency Vehicle Priority System Using MATLAB Simulation

Mohit Jha
Department of Electrical Engineering
Jabalpur Engineering College, Jabalpur, M.P., India
mohitjha_1989@yahoo.com

Shailja Shukla
Department of Computer Science Engineering
Jabalpur Engineering College, Jabalpur, M.P., India
shailja270@gmail.com

*Abstract*—Traffic is the chief puzzle problem which every country faces because of the enhancement in number of vehicles throughout the world, especially in large urban towns. Hence the need arises for simulating and optimizing traffic control algorithms to better accommodate this increasing demand. Fuzzy optimization deals with finding the values of input parameters of a complex simulated system which result in desired output. This paper presents a MATLAB simulation of fuzzy logic traffic controller for controlling flow of traffic in isolated intersections. This controller is based on the waiting time and queue length of vehicles at present green phase and vehicles queue lengths at the other phases. The controller controls the traffic light timings and phase difference to ascertain sebaceous flow of traffic with least waiting time and queue length. In this paper, the isolated intersection model used consists of two alleyways in each approach. Every outlook has different value of queue length and waiting time, systematically, at the intersection. The maximum value of waiting time and vehicle queue length has to be selected by using proximity sensors as inputs to controller for the ameliorate control traffic flow at the intersection. An intelligent traffic model and fuzzy logic traffic controller are developed to evaluate the performance of traffic controller under different pre-defined conditions for oleaginous flow of traffic. Additionally, this fuzzy logic traffic controller has emergency vehicle siren sensors which detect emergency vehicle movement like ambulance, fire brigade, Police Van etc. and gives maximum priority to him and pass preferred signal to it.

## I. INTRODUCTION

Today's conventional controllers, which are developed based on recorded data to ameliorate timing plans are no longer the fanciful Solution to traffic intersections due to varying traffic volumes with respect to time and also increasing numbers of vehicles on road. Traffic controllers which will be able to cogitate equal way of human thinking are designed using Intelligence techniques like fuzzy logic. The main purpose of making new intelligent traffic controllers is that the traffic controllers that have the overall efficiency to accommodate to the present time data from sensors or detectors to perform constant command of interpreter on the signal timing plan for intersections in a network in order to reduce traffic overcrowding which is the main anxiety in traffic flows control hodiernal, at traffic intersections.

Human judgment making and Inference in traffic and carriages are designate by a generally good execution. Even if the judgment makers have unfinished information and key judgment merits are accurately or oracularly as stipulated or not described at all, and the judgment taking goal are ambiguous, the capacity of human judgment building is remarkably. According to [1], traffic intersections that are managed by human operators are still more effective as compared to the traffic responsive control and traditional methods. The older system uses weight as a trigger mechanism Current traffic systems react to motion to trigger the light changes [2].

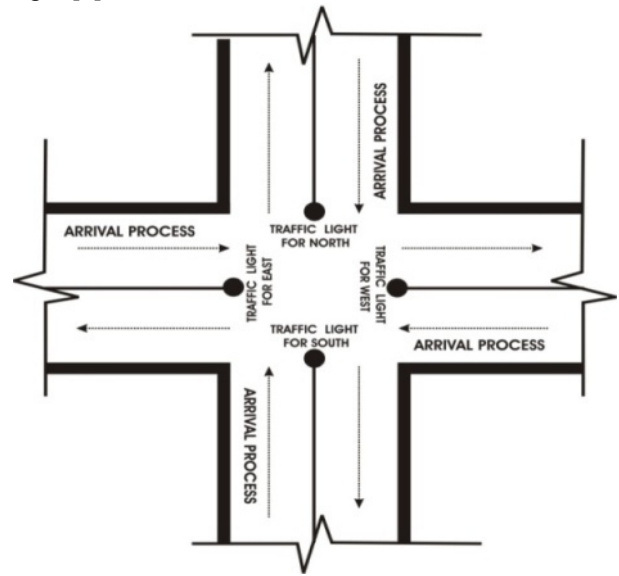

*Figure 1 Segregate Traffic Isolated Intersection Model*

The first step-in-aid of fuzzy logic controller in the history in 1977, which displays preferable execution weigh to vehicle actuated controller for an exclusively intersection have two one-way roadways based on a green time extension principle. From this persuasive work, the main attention for the research has been initiated on petition for fuzzy control methods for intersection control greatly focus at a segregate intersection.

Modern traffic signal controls use highly capable microprocessor based algorithms to control vehicle movements through intersections [3]. The utilization of fuzzy logic controllers in juxtaposition with conventional pre-timed or vehicle-actuated control modes has provided improved traffic manipulation ethically to the usually adopted execution measures as in the case with delays and number of stops. Fuzzy Control is a control method that applies the knowledge of fuzzy mathematics to imitate the human brain's thought. It can recognize and adjudicate the fuzzy phenomenon, and control the system effectively [5]. Fuzzy controllers have perfectly demonstrated dominant in controlling a single traffic intersection, even if the intersection is in certain complex level. In somewhat illustration, even if topical controllers perform nice, there is no clearly warranty that they will continue to do so when the intersections are concatenate with irregular traffic flow. Now, further development took place by accepting fuzzy logic based controllers on traffic signal for two-way single intersection. In Traffic signal intersections, vehicle detection sensors are linked together in order to form an individual closed network [6].

In this research, extensive description on the method used in designing the traffic signal controllers and the overall project development are included. MATLAB is the exclusive software program used in step-in-aid of the whole project. The traffic signal controllers are contemplated using SIMULINK block diagram provided by MATLAB. For fuzzy logic based traffic signal controller system, Mamdani-Type fuzzy inference system (FIS) editor is used to develop fuzzy rules, input and output membership functions. Fuzzy traffic controller will be constitute either using graphical user interface (GUI) tools or working from the command line. In this project, the traffic model is developed using SIMULINK model block diagram and extended with the SimEvent block diagram. Nevertheless, actuated traffic signal controller for isolated intersection is developed in this project. This fuzzy logic traffic controller work separately for emergency vehicles like ambulance, fire-brigade and police van. They give separate time interval for passing an emergency vehicle from intersection according to their movement. The intersection delay time, there have been a variety of achievements in recent years [4]. Lastly, the results from the simulations are shown on waiting time, average delay time and queue length and presence of emergency vehicle in queue as execution index for controlling traffic flow at the intersection.

## II. SEGREGATE TRAFFIC MODEL

The traffic signal controller for segregate intersection is shown in Figure 1 is designed based upon the normal traffic system for two-way single intersection. The segregate traffic intersection model developed in MATLAB using Simulink and SimEvent toolbox is shown in Figure 2.

There are four standpoints in this segregate intersection model with eight total movements and a server traffic light. Each standpoint consists of two campaigns which are one through campaign and one right turn campaign. This model is based upon multiple input single output theory and is constructing based on three main desired concepts in queuing theory which are customers, queue, and servers.

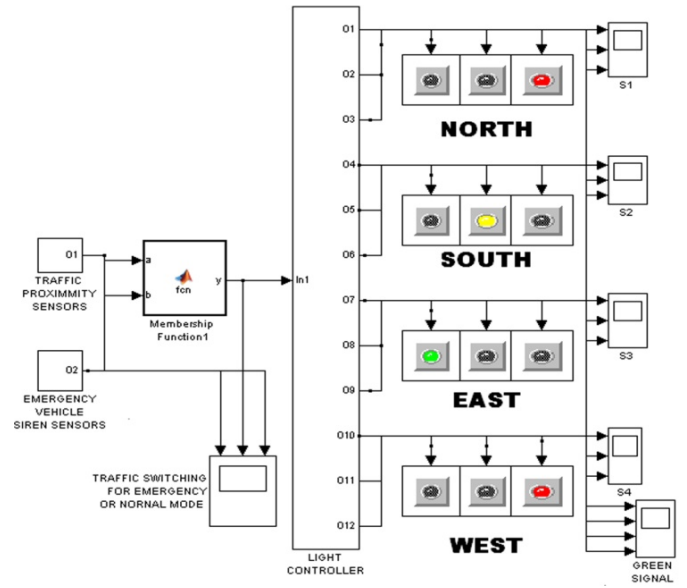

*Figure 2 SIMULINK Block Diagram of Segregate Traffic Intersection Model*

There is one more queue model discipline is applied on all stand points that is first-come-first-out (FIFO). From queuing theory point of view, the vehicles are like customers in this model while services time is the waiting time to get off from intersection. Traffic arrival rate and service times of vehicle at the intersection are independent random variables with Poisson distribution which means that vehicles arrival rate at the intersection is Poisson process with arrival rate λ and the mean of the inter-arrival rate times between vehicles are 1/λ. The arrival vehicle is a Poisson process and the numbers of arrival of vehicles in a system is a Poisson distribution. Function as shown by Equation 1.

$$p\{q_{in}(t) = k\} = \frac{(\lambda \Delta t)^k e^{-\lambda \Delta t}}{k!}$$

(1)

Where, λ is greater than 0 is the arriving rate which is equivalent to the number of arrived vehicles per time period and k=0, 1, 2, ….

## III. DESIGN OF MAMDANI TYPE FUZZY LOGIC TRAFFIC CONTROLLER

For this project Mamdani Type fuzzy logic traffic controller is designed using MATLAB Toolbox. The design has been divided into three stages which are Green Phase stage, next phase stage, switching stage. The design structure of fuzzy logic segregate traffic intersection model controller is shown in Figure 4.

### A. Green Phase Stage

The real time traffic conditions of the green phases are supervised by the Green Phase Stage. Green phase magnitude value for real time is produced by this stage according to the present condition observed by traffic flow using proximity sensors on both side of streets. Fuzzy logic controller block and embedded MATLAB function block that contain C programming codes are the two main blocks of this stage. This stage contain "Fuzzy Controller block" which has one set of Mamdani Type fuzzy inference system which is used to evaluate green signal extension time on real time. In this fuzzy controller there is set of 25 rules and fuzzy inference system this rules takes the value of vehicles waiting time and the vehicle queue length at real time at green phase and creates extension time value as an output. This value is sent to "Embedded MATLAB function" block for assessment. This block contain if-else statement which finds the real probability that the green phase need to extend based on the generated output from the fuzzy inference system and the queue length of the other three phases.

Queue length ($Q_L$) and waiting time ($W_t$) are consumed as the two input variables for fuzzy inference system in traffic controller using proximity sensors which is shown in Figure 3. This system contains input membership function, fuzzy set rules and output membership function. Here, in both input and output membership function Gaussian type membership function is used in place of triangular membership function because traffic does not change linearly in real time. The range of vehicle waiting time is assumed to be 50seconds which is divided into five different ranges: very short (VS), short (S), long (L), very long (VL), and extremely long (EL).

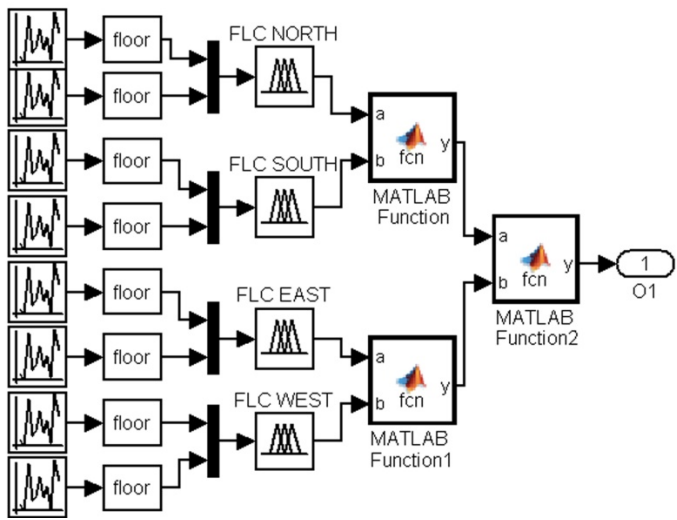

*Figure 3 SIMULINK Block Diagram of Traffic Proximity Sensors*

Each range coincides to a membership function. Also there are five ranges of membership functions in vehicle waiting time ($W_t$). All of these have standard deviation ($\sigma$) of 2 and the constant for Gaussian membership function of very short (VS), short (S), long (L), very long (VL), and extremely long (EL) are of 0seconds, 10 seconds, 20 seconds, 30 seconds, and 40 seconds, respectively.

Similarly, for the vehicle queue length ($Q_L$), the range is assumed to be 0 to 50 vehicles in a lane on each approach at the intersection. The input to a vehicle queue length ($Q_L$) membership function is very short (VS), short (S), long (L), very long (VL), and extremely long (EL). All of these have standard deviation ($\sigma$) of 2 and the constant for Gaussian membership function of very short (VS), short (S), long (L), very long (VL), and extremely long (EL) are of 0vehicles, 10 vehicles, 20 vehicles, 30 vehicles, and 40 vehicles, respectively.

The output fuzzy variable span which means extended time of green signal light is divided into 5 ranges analogous to fuzzy sets: zero (Z), short (S), long (L), very long (VL), and extremely long (EL). All these membership functions are Gaussian type with standard deviation ($\sigma$) of 2 and constant, c which is equals to 2.5.

Fuzzy logic controller is designed with rule base using IF-THEN conditions. Mainly, fuzzy rules system is developed with IF-AND-THEN statements. The fuzzy logic rule base traffic signal controller at segregate intersection is defined is TABLE 1.

TABLE I. FUZZY LOGIC RULE BASE FOR TRAFFIC CONTROLLER

| Rules | Waiting Time ($W_t$) | Queue Length ($Q_L$) | Output |
|---|---|---|---|
| 1. | VS | VS | Z |
| 2. | VS | S | Z |
| 3. | VS | L | S |
| 4. | VS | VL | S |
| 5. | VS | EL | L |
| 6. | S | VS | Z |
| 7. | S | S | S |
| 8. | S | L | S |
| 9. | S | VL | L |
| 10. | S | EL | L |
| 11. | L | VS | S |
| 12. | L | S | S |
| 13. | L | L | L |
| 14. | L | VL | L |
| 15. | L | EL | L |
| 16. | VL | VS | S |
| 17. | VL | S | S |
| 18. | VL | L | L |
| 19. | VL | VL | VL |
| 20 | VL | EL | EL |
| 21. | EL | VS | L |
| 22. | EL | S | L |
| 23. | EL | L | L |
| 24. | EL | VL | VL |
| 25. | EL | EL | EL |

## B. Next Phase Stage

This stage controls the phase order based on the length of vehicle's queue and their extension time of green light from Green Phase Stage. The SIMULINK block diagram of next phase stage is shown in Figure 4. This stage pick one phase for the green signal and it extend the green time of the green phase on the basis of real time traffic condition of the other three phases.

There are four phases in this stage which are Green light on East direction is phase1, Green light on West direction phase2, Green light on South direction phase3, and Green light on North direction phase4. The real time series is controlled by the triggered system. Two output of Next phase stage is connected by switching stage.

## C. Switching Stage

This stage switches current phase to the demanded next phase by output of their previous stage. If any other way has more vehicle queue length than current phase to the next phase basis of output of next phase stage. If the present output of any other phase has more queue length than the queue length of current green signal phase. Then the next phase stage give signal to switching stage to change phase to longer queue. Code for the switching stage is shown in Figure 4.

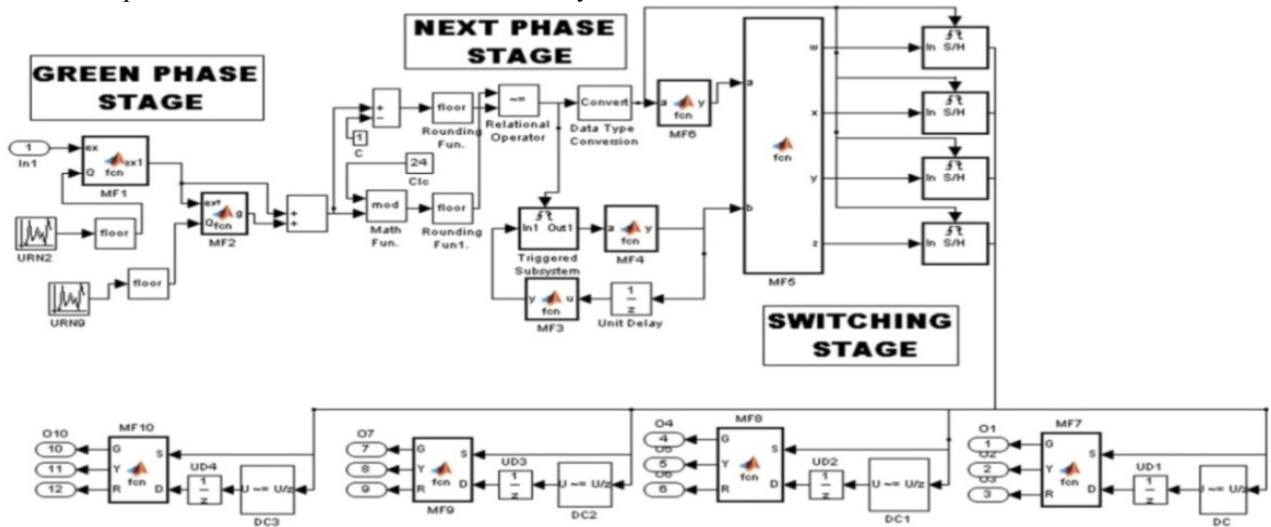

Figure 4 SIMULINK Block Diagram of Light Controller Showing All Three Stages Green Phase, Next Phase And Switching Stage.

## D. Emergency Vehicle Controlling

All the other blocks of the traffic controller is same for the emergency vehicle control system except that an "embedded MATLAB function block" which passes an emergency vehicle queue length and their waiting time to it. This function block has C coding which continuously check for any emergency vehicle siren noise signal and will active only of a particular instant of run time and give maximum priority to emergency vehicle and then after passing emergency vehicle it revert back to their previous stage of real time traffic.

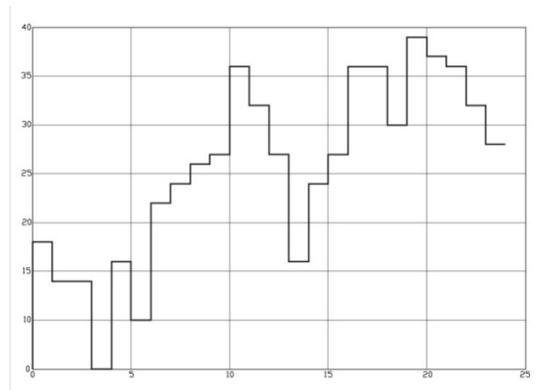

Figure 5(B) Traffic Arrival Process in South Direction

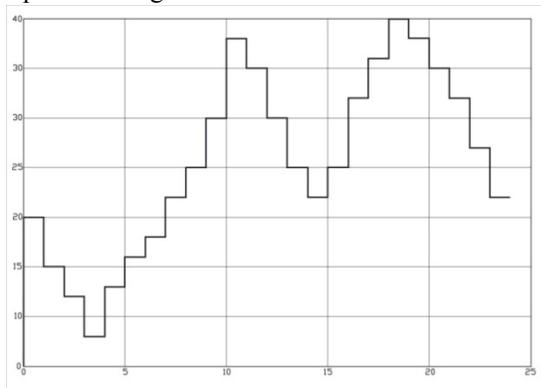

Figure 5(A) Traffic Arrival Process in North Direction

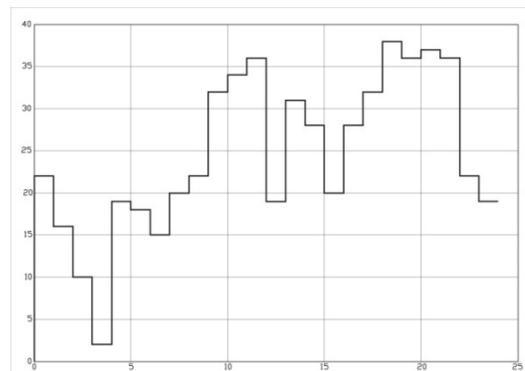

Figure 5(C) Traffic Arrival Process in East Direction

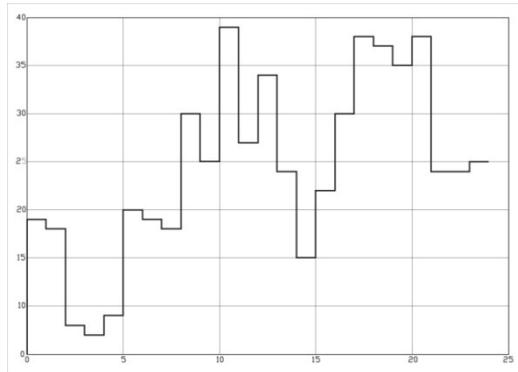

*Figure 5(D) Traffic Arrival Process in West Direction*

## IV. CONCLUSIONS

In this paper, the traffic model and traffic controller are develop using MATLAB software. This paper is based on queuing theory model of multiple–input-single-output. The traffic model is simple to construct using SIMULINK model and SimEvent toolbox in MATLAB. The traffic controller is developed using fuzzy inference system method in MATLAB.

To test the effectiveness of this controller here four different recorded data is considered shown in Figure 5(A), 5(B), 5(C) and 5(D). Also, use certain emergency vehicle data and test over run time and check the output graph both for real traffic case and an emergency vehicle case.

Simulation results of green phase switching shown in Figure 6(A), 6(B), 6(C) and 6(D) proves that fuzzy logic traffic controller is superior to any classical or timing control methods. Fuzzy control system scheme avoids the vehicles waiting in crossing as much as possible, mitigates the traffic congestion effectively, improves the intersection vehicle crossing capacity and realizes the intelligent control of traffic lights. This system is also works well intelligently for an emergency vehicle case Traffic movement shown in Figure 7(A) Normal Mode, 7 (B) an Emergency Vehicle Arrival Mode and 7(C). Intelligent crossing system is the next generation transportation system, as an important part of intelligent traffic light control system has important significance and potential applications in a whole world.

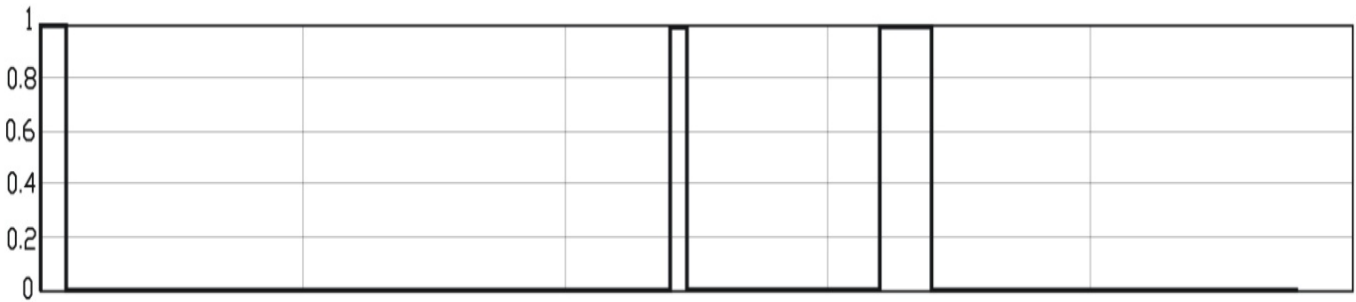

*Figure 6(A) Green Signal Switching in North Directions*

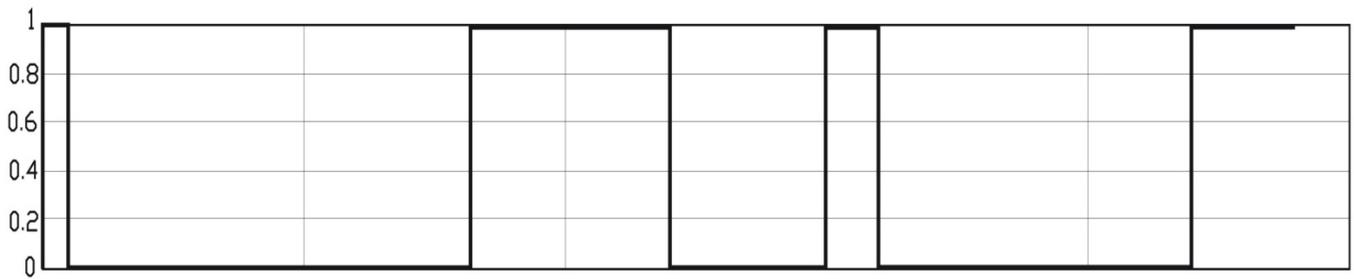

*Figure 6(B) Green Signal Switching in South Directions*

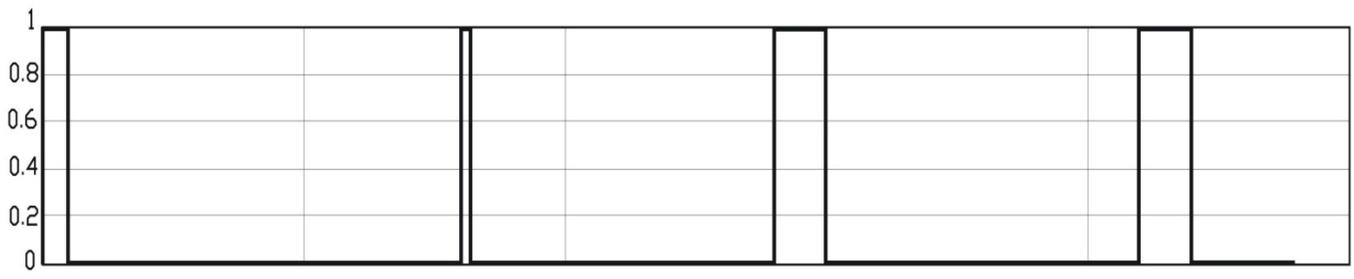

*Figure 6(C) Green Signal Switching in East Directions*

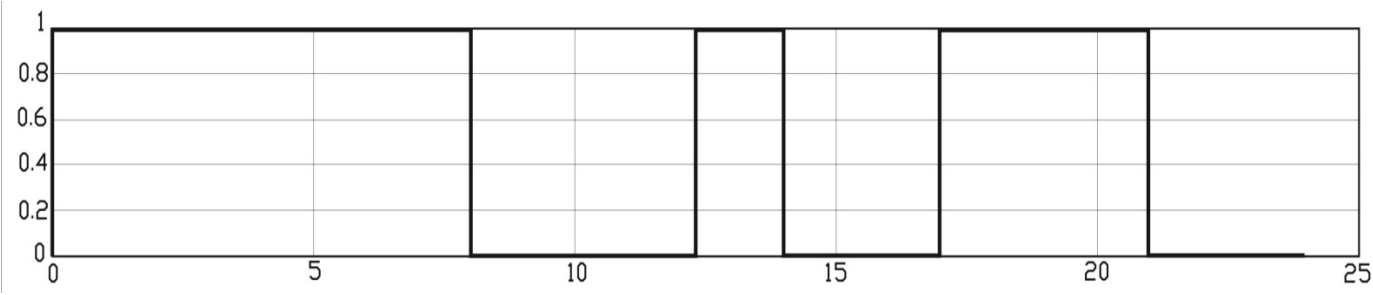
*Figure 6(D) Green Signal Switching in West Directions*

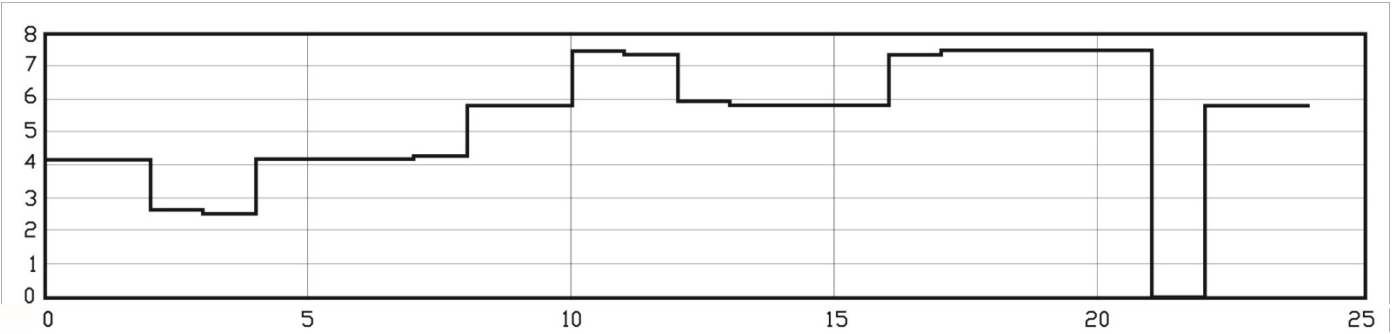
*Figure 7(A) Traffic Movements in Normal Mode*

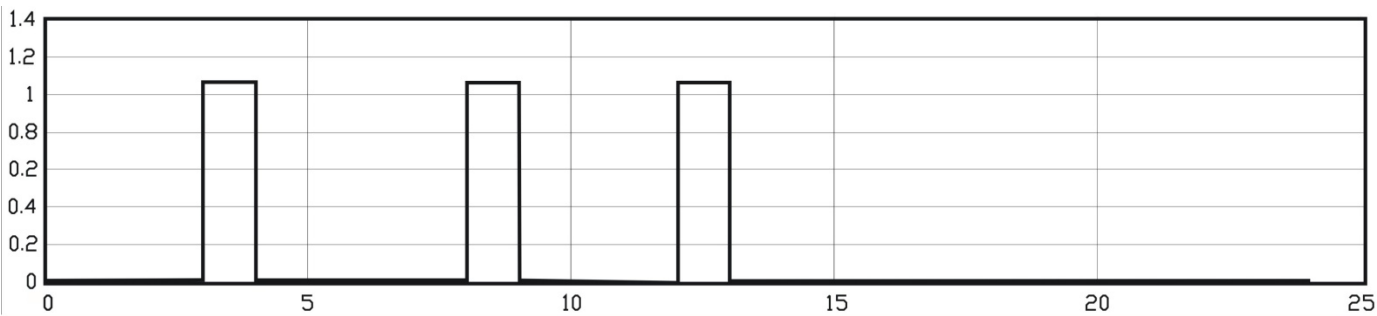
*Figure 7(B) Traffic Movements in Emergency Vehicle Arrival Mode*

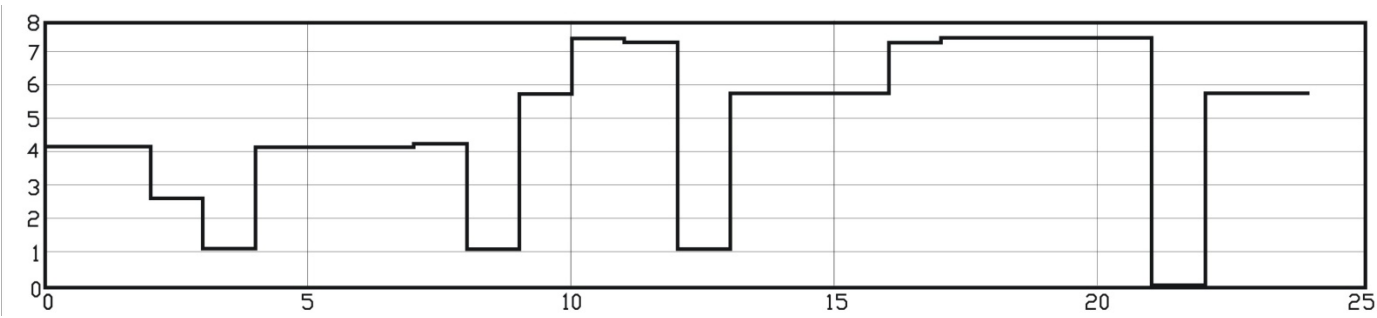
*Figure 7(C) Traffic Movements Final Concluded Output*